\documentclass[10pt,twocolumn,letterpaper]{article}

\usepackage{cvpr}              

\usepackage{times}
\usepackage{epsfig}
\usepackage{graphicx}
\usepackage{amsmath}
\usepackage{amssymb}
\usepackage{booktabs}
\usepackage{amsfonts}       
\usepackage{nicefrac}       
\usepackage{microtype}      
\usepackage{xcolor}         
\usepackage{multicol}
\usepackage{multirow}
\usepackage{pifont}
\usepackage{graphicx}

\usepackage{subcaption}
\usepackage{booktabs}
\usepackage{amsmath}
\usepackage{relsize}
\usepackage[accsupp]{axessibility}

\usepackage[pagebackref,breaklinks,colorlinks]{hyperref}

\usepackage[capitalize]{cleveref}
\crefname{section}{Sec.}{Secs.}
\Crefname{section}{Section}{Sections}
\Crefname{table}{Table}{Tables}
\crefname{table}{Tab.}{Tabs.}

\newcommand{\Te}{\emph{Teacher}}
\newcommand{\St}{\emph{Student}}
\newcommand\blfootnote[1]{%
  \begingroup
  \renewcommand\thefootnote{}\footnote{#1}%
  \addtocounter{footnote}{-1}%
  \endgroup
}

\def\cvprPaperID{8} 
\def\confName{CVPR}
\def\confYear{2022}

\begin{document}
\title{Holistic Approach to Measure Sample-level Adversarial Vulnerability \\ and its Utility in Building Trustworthy Systems}
\author{Gaurav Kumar Nayak\thanks{\scriptsize{equal contribution.}}\quad
Ruchit Rawal\footnotemark[1]\quad 
Rohit Lal\footnotemark[1]\quad 
Himanshu Patil $^{1}$\quad
Anirban Chakraborty\\
Indian Institute of Science, Bangalore, India\\
{\tt\small\{gauravnayak, ruchitrawal,  rohitlal, anirban\}@iisc.ac.in}\quad $^{1}$\tt\small{hipatil1998@gmail.com}
}
\maketitle
\vspace{-0.3in}
\begin{abstract}
Adversarial attack perturbs an image with an imperceptible noise, leading to incorrect model prediction. Recently, a few works showed inherent bias associated with such attack (robustness bias), where certain subgroups in a dataset (e.g. based on class, gender, etc.) are less robust than others. This bias not only persists even after adversarial training, but often results in severe performance discrepancies across these subgroups. Existing works characterize the subgroup’s robustness bias by only checking individual sample’s proximity to the decision boundary. In this work, we argue that this measure alone is not sufficient and validate our argument via extensive experimental analysis. It has been observed that adversarial attacks often corrupt the high-frequency components of the input image. We, therefore, propose a holistic approach for quantifying adversarial vulnerability of a sample by combining these different perspectives, i.e., degree of model’s reliance on high-frequency features
and the (conventional) sample-distance to the decision boundary. We demonstrate that by reliably estimating adversarial vulnerability at the sample level using the proposed holistic metric, it is possible to develop a trustworthy system where humans can be alerted about the incoming samples that are highly likely to be misclassified at test time. This is achieved with better precision when our holistic metric is used over individual measures. To further corroborate the utility of the proposed holistic approach, we perform knowledge distillation in a limited-sample setting. We observe that the student network trained with the subset of samples selected using our combined metric performs better than both the competing baselines, viz., where samples are selected randomly or based on their distances to the decision boundary. \blfootnote{{\textit{\scriptsize{Webpage}}: \scriptsize{\url{https://sites.google.com/view/sample-adv-trustworthy/}}
}} 
\end{abstract}
\vspace{-0.05in}
\section{Introduction}
\label{sec:intro}
Deep neural networks (DNNs) are becoming increasingly ubiquitous across a plethora of real-world applications such as object detection \cite{zhao2019object,wu2020recent}, speech recognition \cite{noda2015audio,kamath2019deep}, remote sensing \cite{zhu2017deep, ma2019deep}, etc. However, these models are extremely brittle, as they yield incorrect predictions on samples corrupted by carefully crafted perturbations that are imperceptible to humans, popularly known as adversarial samples \cite{szegedy2013intriguing,goodfellow2014explaining,madry2017towards}. Vulnerability towards adversarial examples (adversarial vulnerability) in state-of-the-art DNNs is particularly worrisome in safety-critical applications such as medical imaging \cite{barucci2020adversarial, paschali2018generalizability}, facial recognition \cite{yang2021attacks,vakhshiteh2021adversarial}, self-driving cars \cite{morgulis2019fooling,eykholt2018robust}, etc. For 
example, adversarially manipulated traffic signs can be misclassified by a self-driving car leading to severe implications like loss of human life.

Recently, researchers have demonstrated \cite{nanda2021fairness, papernot2016limitations} that the notion of adversarial robustness also coincides with fairness considerations as certain subgroups (often corresponding to sensitive attributes like gender, race etc.) in a dataset are less robust in comparison to the rest of the data and thus are harmed disproportionately in case of an adversarial attack.  A subgroup can broadly be defined as categorization of input-space into disjoint partitions based on certain criteria of interest such as class labels, sensitive attributes, etc. Nanda \etal \cite{nanda2021fairness} in their work, formulated the phenomena of different levels of robustness exhibited by different subgroups as ‘robustness bias’. Moreover, robustness bias is present (and sometimes even amplified) \cite{pmlr2021benz, pmlr21xu} after supposedly making the model ‘robust’ using state-of-the-art adversarial defense methods like adversarial training. Ideally, a fair model should ensure each (and any) subgroup in the dataset is equally robust. For example, in a face recognition system, the model should perform equally well across different ethnic subgroups.

\begin{figure*}[htp]
    \centering
    \includegraphics[width=0.98\linewidth]{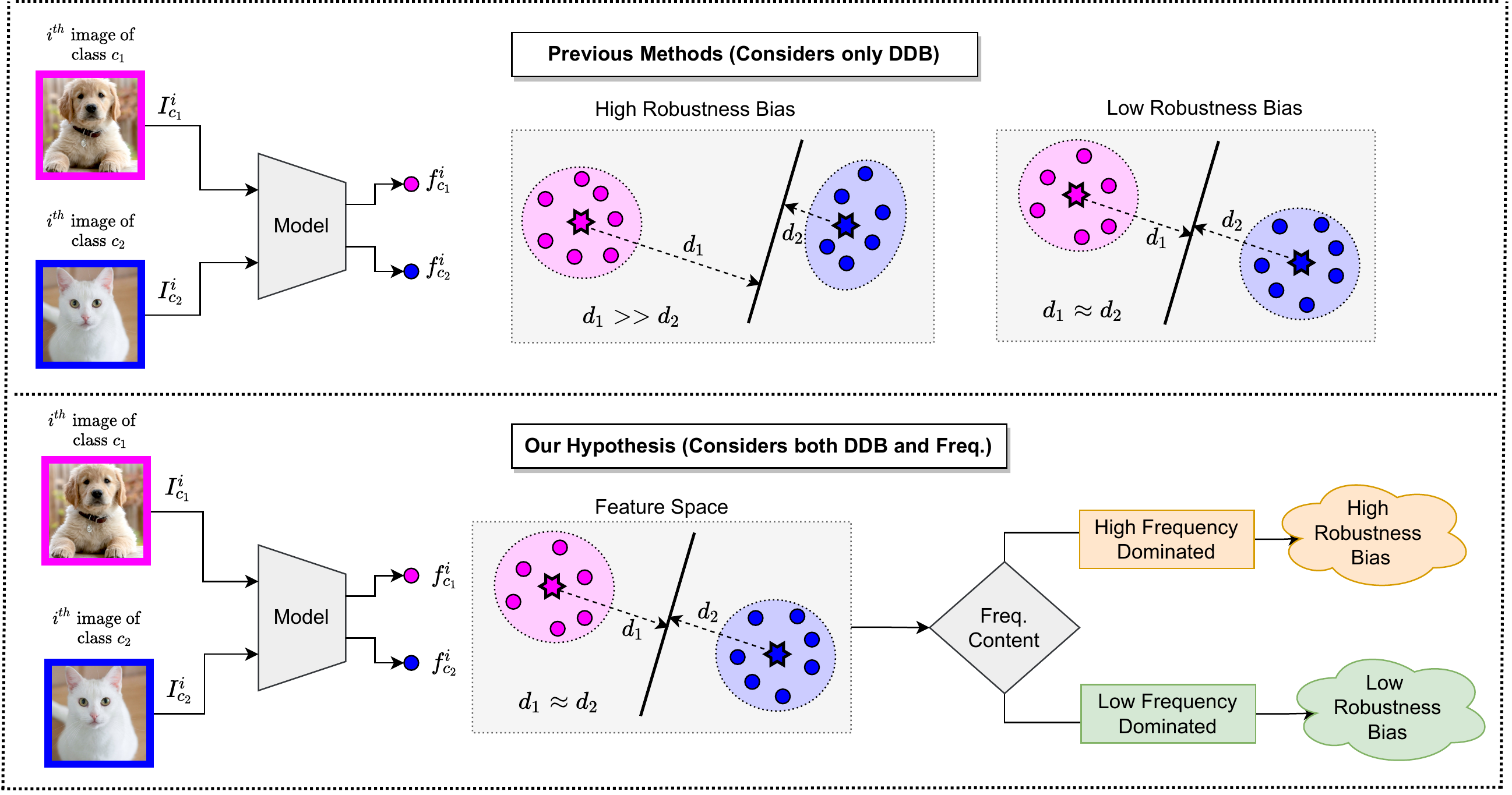}
    \caption{Comparison of existing methods vs our methods: Existing methods characterise robustness bias only via distance to decision boundary (DDB) as the only factor for describing adversarial vulnerability whereas our objective is to incorporate mutlitple perspectives which considers multiple factors (frequency and DDB). Subgroups (shown in pink and blue colour) lying at a similar distance to decision boundary has lower robustness bias compared to subgroups having different DDB (one subgroup lies closer while the other lies farther). Our hypothesis is that even samples lying at similar DDB can have high robustness bias if they are dominated by high frequencies.}
    \label{fig:compare_methods}
\end{figure*}
Accurately measuring the adversarial vulnerability of a sample is the first step towards designing generalizable defense framework that can mitigate robustness bias, as the notion of subgroups may vary significantly depending on the tasks and datasets. Traditionally, this is achieved by using distance to decision boundary (DDB) as a proxy to quantify sample robustness 
i.e. samples that are farther away from decision boundary are assumed to have higher robustness and vice-versa. Proximity to the decision boundary is an intuitive measure as samples lying closer to the decision boundary would naturally be more prone to cross the decision boundary after perturbation, eventually leading to misclassification. 
However, in our analysis (Sec.~\ref{sec:DDB_insufficient}), we find that the samples with similar DDB values can have different adversarial vulnerabilities, i.e., such instances can require varying minimal steps (by an iterative adversarial attack) to fool the model. We observe this trend to be consistent across adversarial attacks. 
Furthermore, we also provide 
explanations 
about these observations (Sec.~\ref{sec:reasons_DDB}). Hence, unlike \cite{nanda2021fairness}, we argue based on experimental observations that DDB (although important) alone cannot be the only factor to 
measure adversarial vulnerability, motivating us to propose a holistic view for estimating adversarial vulnerability by (also) taking other factors into consideration.

Another perspective to quantify adversarial robustness is based on a trained model's reliance on high-frequency features in the dataset. While humans primarily rely on low-frequency (LF) features \cite{wang2020towards}, the DNNs, on the contrary, not only focuses on LF features but can also extract ‘useful’ predictive features from the high-frequency (HF) components in the data to maximize their performance \cite{wang2020high}. Interestingly, many state-of-the-art adversarial attacks primarily perturb the HF component of a sample in order to force erroneous predictions. Thus, the reliance of DNNs on HF features for better generalization makes them highly vulnerable to adversarial attacks \cite{wang2020towards}. These observations indicate that robustness to adversarial attacks is not only dependent on the distance to decision boundary but also on the nature of the features learned (discussed in Sec.~\ref{sec:another_perspective}).

This work aims to investigate the adversarial vulnerability of a sample from two distinct perspectives, i.e., a) proximity to decision boundary, b) reliance on high-frequency features. Our objective is to provide a holistic estimation, as only relying on proximity to decision boundary might give a false sense of robustness. The difference between existing works and ours is also shown in Figure~\ref{fig:compare_methods}. 

Based on the above perspectives, we propose the idea of a trustworthy systems (Figure~\ref{fig:trust_system}) where the trained models would also have provision to yield trust score besides regular class label predictions. The trust score (quantified using both DDB and HF factor) indicates the reliability of a model’s prediction. A test sample is clustered either into `Trust' or `Non-Trust' cluster based on its trust score. The predictions are reliable only when the sample falls in the Trust cluster. Otherwise, samples in the Non-Trust cluster can be surfaced to a domain-expert for further inspection and annotation. 

An analysis of multiple factors for quantifying adversarial vulnerability can also be utilized for different applications such as knowledge distillation. Specifically, a student network can be trained efficiently in a limited-data setting by selecting the samples based on adversarial vulnerability factors (viz., distance to decision boundary and reliance on high-frequency). We empirically observe better performance by selecting samples based on our proposed criterion, compared to random-baseline and conventional methods (details in Sec.~\ref{sec:KD}).

Our overall contributions are summarized as follows:
\begin{itemize}
\item We show that the traditional notion of quantifying adversarial vulnerability, i.e., `proximity to the decision boundary', is insufficient, leading to a false sense of robustness.  We verify it across multiple adversarial attacks and architectures, and provide suitable justification for the same.
\item We incorporate multiple perspectives (distance to decision boundary and reliance on high frequency) 
to characterize robustness bias and thereby provide a more holistic view to estimate 
adversarial vulnerability. This is still unexplored to the best of our knowledge.
\item We build a trustworthy system that predicts the sample-level trust scores based on the multiple factors of 
sample vulnerability. 
We further demonstrate that our proposed system alerts the human domain experts when the samples are highly likely to be misclassified (i.e. non-trustable) with better precision than the system designed by using individual factors only.
\item As a case study on the task of knowledge distillation under limited data setting, 
we show the utility of multiple perspectives of adversarial vulnerability in sample selection for composing better transfer set. We obtain improvements in \St{}'s performance across different low-data settings against multiple baselines and this trend remains consistent even when the hyperparameters (e.g. temperature in distillation loss) are varied. 
\end{itemize}

\section{Related Works}
\label{sec:related_works}
Several hypothesis for explaining the existence of adversarial samples have been proposed in recent years \cite{ilyas2019adversarial, wang2020towards, geirhos2018imagenet, fawzi2018adversarial, gilmer2018adversarial}. Most prominently, researchers~\cite{ilyas2019adversarial} have attributed the presence of adversarial samples to useful yet brittle i.e. non-robust features in the data that can be easily latched on to, by DNNs. On similar lines, Wang \etal~\cite{wang2020high} demonstrated that a DNN can learn well-generalizable features from high-frequency components in the data. These high-frequency components are shown to be most prone to perturbations (i.e. higher adversarial vulnerability) in case of state-of-the-art adversarial attacks~\cite{wang2020towards}.

Fairness approaches aim to ensure that no subgroup is disproportionately harmed (or benefited) due to decisions taken by an automated-decision making system~\cite{hooker2020characterising, zafar2019fairness, morales2020sensitivenets}. Recently, Nanda \etal~\cite{nanda2021fairness} demonstrated that adversarial robustness is closely connected to the notion of fairness as different subgroups possess different levels of robustness (robustness bias) leading to unfair scenarios in downstream applications. Furthermore, robustness bias persists~\cite{pmlr2021benz, pmlr21xu} even after making the model robust using adversarial defenses like adversarial training.

Traditionally, distance to decision boundary (of a sample) is used as a proxy for measuring it's adversarial vulnerability~\cite{nanda2021fairness}. However, we argue that such a measure is inadequate as a sample could be adversarially vulnerable due to other factors as well. Thus, our objective is to holistically estimate such adversarial vulnerability by combining multiple perspectives 
(namely proximity to decision boundary and reliance on high-frequency features) and further show it as a better measure than the individual factors alone. 
\vspace{-0.1in}
\section{Preliminaries}
\label{sec:preliminaries}
\textbf{Notations}: The model $M$ is trained on a labelled dataset $\mathcal{D} = \{(x^{i}, y^{i})\}_{i=1}^{N}$ using a standard cross entropy loss ($\mathcal{L}_{ce}$). The dataset $\mathcal{D}$ contains images from $p$ different classes i.e. the image set $I= \{I_{ck}\}_{k=1}^{p}$ and $I_{ck}$ represents set of images from $k^{th}$ class. Each class contains equal amount of samples i.e. $N/p$. $M(x^{i})$ denotes the logits and the class prediction by the model $M$ on the $i^{th}$ input ($x^{i}$) is represented by $argmax(softmax(M(x^{i})))$. 

$A_{adv}$ contains a set of $s$ different adversarial attacks i.e. $A_{adv} = \{A^{j}_{adv}\}_{j=1}^{s}$. An $i^{th}$ adversarial sample (i.e. $\hat{x}^{i}$) is obtained by perturbing the sample ($x^{i}$) with an objective to fool the network $M$. 

The Discrete Cosine Transform and its inverse operations are denoted by $DCT$ and $IDCT$ respectively. An $i^{th}$ spatial sample ($x^{i}$) has $f^{i}$ as its representation in frequency space.

The model $M$ is made trustworthy where the trust score (denoted by $T$) of a sample is computed at test time and humans are alerted if the sample falls in the non-trustable cluster.

\textbf{Threat model}: The perturbations added via an adversarial attack $A^{j}_{adv} \in A_{adv}$ (also called adversarial noise) is constrained in a $L_{p}$ norm ball to make them human imperceptible i.e. $\left\lVert \hat{x}^{i} - x^{i} \right\rVert_{p} \le \epsilon$.  $L_{\infty}$ and $L_{2}$ are the  two popular threat models. Besides the constraint on $\delta$, the other objective is to enforce the model $M$ to change its prediction on adversarial sample $\hat{x}^{i}$ i.e. $argmax(softmax(M(\hat{x}^{i}))) \neq argmax(softmax(M(x^{i})))$. To achieve these, different optimization procedures are followed leading to different adversarial attacks such as PGD~\cite{madry2017towards} and Deepfool~\cite{moosavi2016deepfool}.

\textbf{Adversarial Training:} At every iteration, the batch of samples from dataset $\mathcal{D}$ is augmented with corresponding adversarial samples, and together they are used to optimize the model parameters by minimizing the loss to obtain a robust model ($\hat{M}$). In PGD adversarial training, the loss taken is the standard cross-entropy loss.

\begin{figure*}[htp]
    \centering
    \begin{subfigure}{0.46\linewidth}
        \centering
        \includegraphics[width=\linewidth]{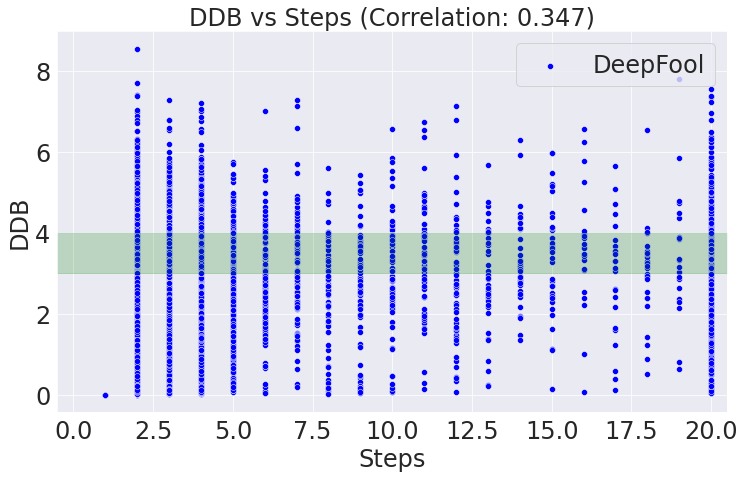}
    \end{subfigure}
    \begin{subfigure}{0.46\linewidth}
        \centering
        \includegraphics[width=\linewidth]{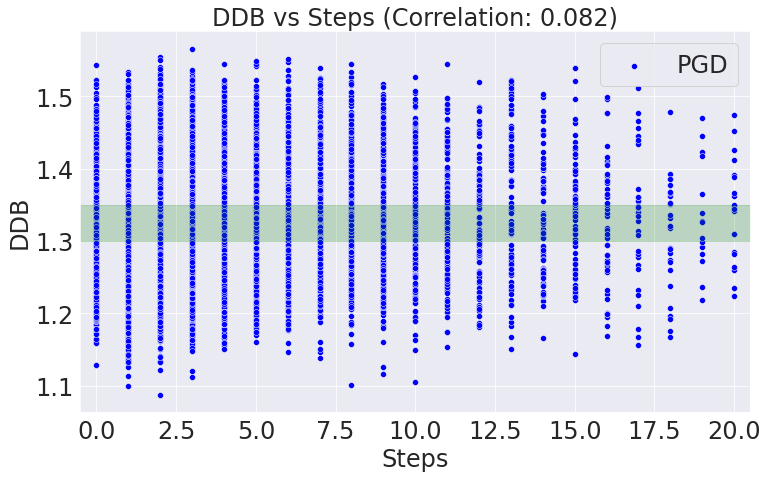}
    \end{subfigure}
    \caption{Plots of Distance to Decision Boundary (DDB) vs Steps to attack (Steps) for different attacks, DeepFool (left) and PGD (right) for CIFAR10 dataset on ResNet18 architecture. Even at a similar distance to decision boundary, number of steps required for the adverserial attack may vary from sample to sample.}
    \label{fig:ddb}
\end{figure*}
\textbf{Distance to decision boundary (DDB)}: The accurate estimation of the distance from the nearest decision boundary $d_{f}(x^i)$ of a trained classifier ($M$ or $\hat{M}$), for any input sample $x^i \in \mathcal{D}$, is an extremely challenging task. The strategy adopted by Nanda \textit{et al.}~\cite{nanda2021fairness} for computing $d_{f}(x^i)$ is to perturb $x^i$ with the aim of transporting it in the input space to a different category than the original. This can be precisely achieved by a standard adversarial attack setup, wherein an adversarial attack $A^{j}_{adv} \in A_{adv}$ computes the adversarial sample $\hat{x}^i$. The difference in predicted labels for $x^i$ and $\hat{x}^i$ by the model indicates that the decision boundary is atmost $\delta^{i} = \left\lVert \hat{x}^{i} - x^{i} \right\rVert_{2}$ distance away from $x^{i}$, establishing $\delta^{i}$ as a reliable estimate of  $d_{f}(x^{i})$.

\textbf{Discrete Cosine Transform (\textit{DCT})}: \textit{DCT} is used to transform the sample from spatial domain to frequency domain i.e. an $i^{th}$ sample of $\mathcal{D}$ ($x^{i}$) gets converted to $f^{i}$. Unlike Discrete Fourier Transform which outputs a complex signal, $DCT$ outputs only a real-valued signal which makes the analysis of images in the frequency domain much easier. In the frequency transformed image $f^{i}$, the top left corner represents the low-frequency content of the image, and the bottom right corner represents the high-frequency contents.  In order to retrieve the spatial sample ($x^{i}$) back from its corresponding frequency domain representation in ($f^{i}$), we use the Inverse Discrete Cosine Transform (\textit{IDCT}).

\textbf{Flipping Frequency}: In order to quantify the reliance of a sample on high-frequencies, we progressively remove high-frequency bands until the model changes it's prediction (w.r.t the prediction on original sample). For instance, if model predicts correctly on the sample containing frequency content till $(k+1)^{th}$ band, and mis-classifies on $k^{th}$ band, this implies that model is dependent on $(k+1)^{th}$ band to make the correct prediction. Hence, if the flipping frequency band ($k+1$ in this case) is high it means that the model relies more on high-frequency content, if flipping frequency is low it means that the model relies on low-frequency.



\section{DDB is insufficient to characterize robustness}
\label{sec:DDB_insufficient}
Fig.~\ref{fig:ddb} contains the plots for model $M$ (Resnet-$18$) trained on CIFAR-10 dataset. The x-axis represents the minimal steps required by the adversarial attack to change the model prediction. Here, we also varied the adversarial attacks ($s=2$ i.e. $A_{adv} = \{PGD, Deepfool\}$) which are used to compute steps to attack 
for each sample in the dataset. The y-axis represents the DDB values for all the samples i.e. $d_{f}(x^{i})_{i=1}^{N}$. We can observe that samples having similar DDB values can require a varying number of attack steps (an instance is highlighted using a rectangular box in green color). This observation is consistent across different attacks.

More the number of steps taken by the adversarial attack to fool a network, the harder the sample is to attack. In other words, more steps imply less adversarial vulnerability (more robust) and vice versa. This along with preceding discussion also implies that the samples with similar DDB (either in close or far regions), can have different adversarial vulnerabilities. This contradicts the traditional assumption that samples that are far from the decision boundary would always be less vulnerable to adversarial attacks and samples with same DDB cannot have varying vulnerabilities. Hence, in contrast to that assumption, even samples far off from decision boundary can require fewer steps to attack (less vulnerable). Moreover, the correlation between DDB and steps to attack is low. Thus, it suggests that DDB is not a sufficient factor to characterize the robustness/vulnerability associated with a sample.
\begin{figure}[htp]
    \centering
    \includegraphics[width=0.8\linewidth]{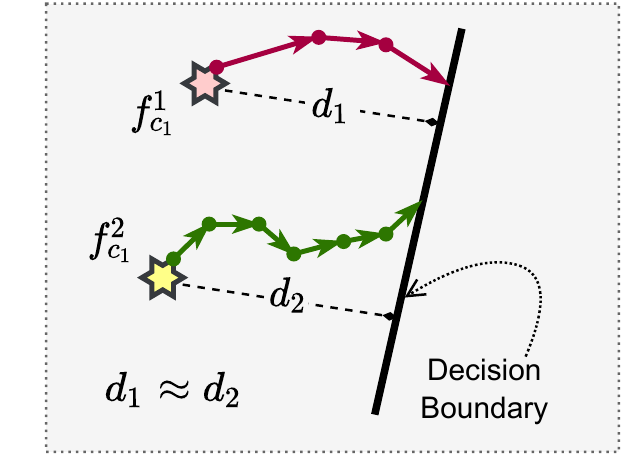}
    \caption{Samples lying to same distance to decision boundary ($f^1_{c_1}$ and $f^2_{c_1}$) can follow different number of optimization steps while generating adversarial samples through adversarial attack (as the path can be highly non-linear), Hence such samples can have diff adv vulnerabilities even after having similar DDB values. }
    \label{fig:optimization}
\end{figure}
\vspace{-0.1in}
\section{
Explanation for DDB as an unreliable estimate of adversarial vulnerability}
\label{sec:reasons_DDB}
To compute DDB of a sample $x^i$ (i.e. $d_{f}(x^i)$), $\delta^{i} = \left\lVert \hat{x}^{i} - x^{i} \right\rVert_{2}$ is chosen as an estimate of DDB. (Refer to Sec.~\ref{sec:preliminaries}). However, as shown in Figure~\ref{fig:optimization}, samples having similar $\delta$ values can take different optimization paths while 
estimating the minimal steps to flip the label via adversarial attack. Hence, samples with similar $\delta$ can have different counts of steps to attack, resulting in different adversarial vulnerabilities. We also note that DDB can become a good estimate if the sample path during optimization is linear. But in general, this is rather strong and improbable assumption and the 
optimization path is highly likely to be non-linear. So, intuitively it may seem that DDB may characterize robustness of a sample, but practically it is not a reliable estimate for deep networks.

\section{Different Perspective: Model Reliance on High Frequency Features}
\label{sec:another_perspective}
In contrast to humans, DNNs heavily rely on high-frequency (HF) features in the data, which makes them susceptible to adversarial attacks \cite{wang2020towards, wang2020high}. Along similar lines, Xu \etal~\cite{pmlr21xu}  demonstrated that the vulnerability of certain classes (that are inherently more-vulnerable/less-robust to adversarial perturbations) is amplified after adversarial training. We hypothesize that such robust DNN must be relying more on high-frequency features for classes most vulnerable to adversarial attacks. 

\begin{figure}[htp]
    \centering
    \includegraphics[width=\linewidth]{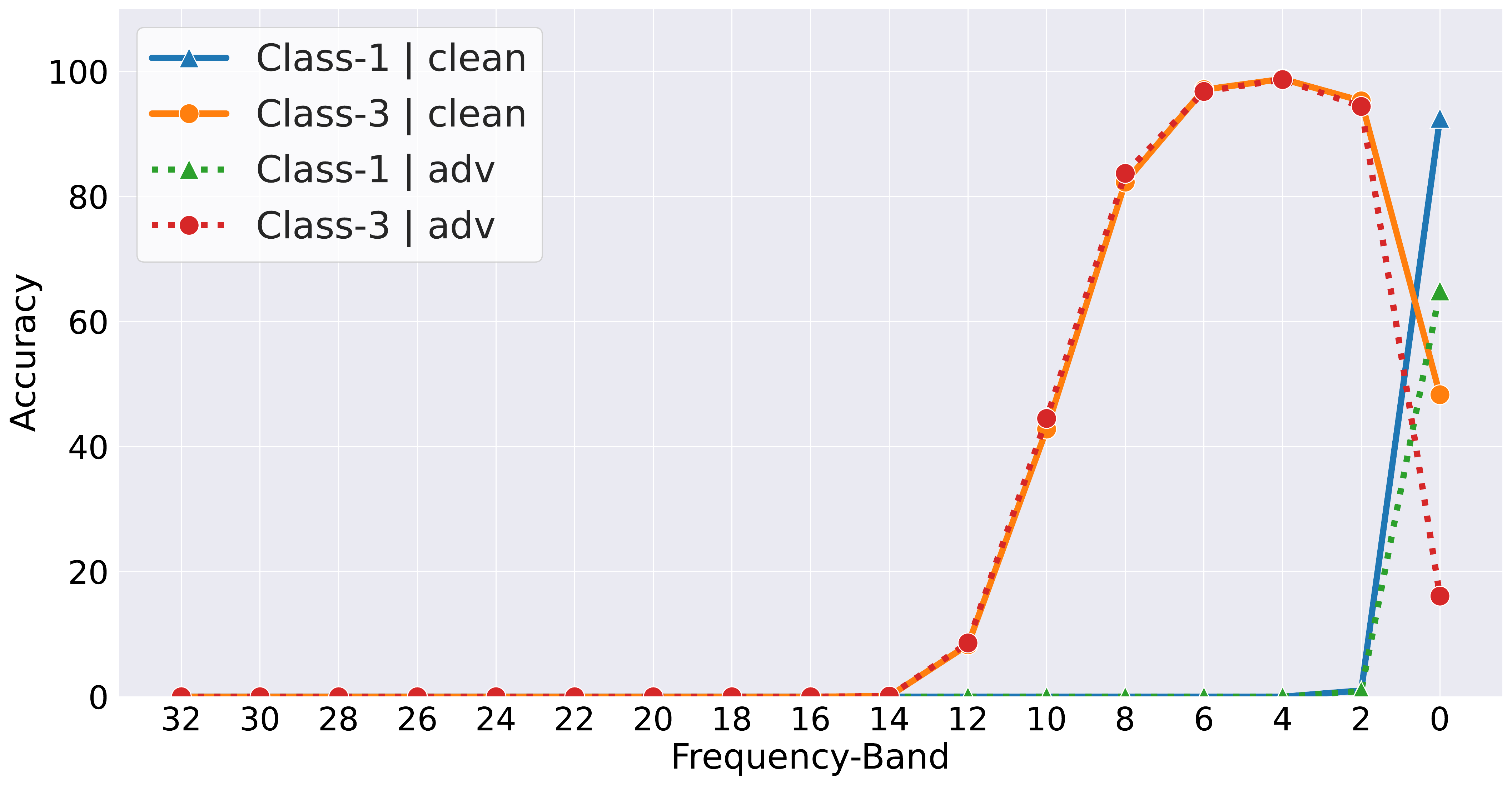}
    \caption{Evaluating robust ResNet-$18$'s performance on multiple frequency bands by gradually adding high-frequency content from the maximum frequency component to the lowest.}
    \label{fig:high_f_acc_comb}
\end{figure}

We empirically validate our hypothesis by calculating the average high-frequency bands required by the model in order to have predictions identical to those original samples for the most and least vulnerable classes (refer Table~\ref{tab:hf-req}). For instance, class-3 (most-vulnerable class) of the CIFAR-10 dataset, on average, requires $32$ (i.e. the maximum frequency band - $k_{max}$) to $\approx9$ 
HF bands to have predictions identical to those on original (full-frequency spectrum) samples. On the contrary, class-1 (least-vulnerable class) requires $32$ to $\approx1$
. Thus, the highly-vulnerable class primarily requires high-frequency information, whereas the least-vulnerable class relies more on low-frequency content. 
We perform additional analysis (refer Figure~\ref{fig:high_f_acc_comb}) to evaluate a robust model’s performance on multiple high-frequency bands, i.e., we gradually add frequency bands, starting from the maximum frequency component (i.e. $k_{max} = 32$) to the lowest ($k_{min} = 0$). In line with our previous observations, we note that the robust model achieves fairly decent performance on the class-3 (most vulnerable) with only high-frequency content, compared to class-1 (least vulnerable), which also leverages low-frequency information to predict accurately.

\begin{table}[htp]
\centering
\scalebox{0.95}{
\begin{tabular}{|c|c|c|c|}
\hline
\textbf{Class} &
  \textbf{\begin{tabular}[c]{@{}c@{}}Clean \\ Accuracy\end{tabular}} &
  \textbf{\begin{tabular}[c]{@{}c@{}}PGD\\ Accuracy\end{tabular}} &
  \textbf{\begin{tabular}[c]{@{}c@{}}Avg. HF \\ Band Req.\end{tabular}} \\ \hline \hline
\begin{tabular}[c]{@{}c@{}}Class-1 \\ (Least Vulnerable)\end{tabular} &
  92.50 &
  64.80 &
  1.06 \\ \hline
\begin{tabular}[c]{@{}c@{}}Class-3 \\ (Most Vulnerable)\end{tabular} &
  48.30 &
  16.40 &
  9.44 \\ \hline
\end{tabular}
}
\caption{Performance of least-vulnerable and most-vulnerable class for a Robust ResNet-18 model on the CIFAR-$10$ dataset. There is significant discrepancy in clean and adversarial performance of both the classes. The more-vulnerable class relies on high-frequency information to predict consistently (w.r.t original samples), whereas the least-vulnerable class relies on low-frequency content.}
\label{tab:hf-req}
\end{table}
\vspace{-0.1in}
\vspace{-0.15in}
\section{Combined Study of DDB and Frequency}
In the previous sections (Sec.~\ref{sec:DDB_insufficient},~\ref{sec:reasons_DDB} and~\ref{sec:another_perspective}), we motivated the use of different perspectives such as DDB and Frequency towards vulnerability analysis. In order to come up with a more holistic measure to quantify adversarial vulnerability of a sample, we introduce a combined score ($T$) that appropriately combines the contribution from both the factors. 

We first obtain the DDB $d_{f}(x^i)$ and Flipping Frequency $F(x^i)$ $\forall i \in \{1, N\}$ (Refer Sec.~\ref{sec:preliminaries}). Since $d_{f}(x^i)$ and $F(x^i)$ can be of different range depending on datasets and other various factors, we therefore normalise both the metrics to a common range $[0, 1]$. We call these normalized values as Normalised DDB and Normalised Flipping Frequency respectively. 

\textbf{Normalised Distance to Decision Boundary}: To obtain Normalised DDB or $\hat{d}_{f}(x^i)$ we simply scale the DDB value between $0$ to $1$ by using Eq. \ref{eq:norm_ddb}.
\begin{equation}\label{eq:norm_ddb}
   \hat{d}_f(x^i) = \frac{d_{f}(x^i) - min(D_{f}(x))}{max(D_{f}(x)) - min(D_{f}(x))} 
\end{equation}
Here $min(D_{f}(x))$ and $max(D_{f}(x))$ denotes minimum and maximum DDB values respectively, where $D_{f}(x) = \{d_{f}(x^1), d_{f}(x^2), \cdots, d_{f}(x^N)\}$ for a dataset. 
Sample farthest from the decision boundary will have $\hat{d_f}(x^i)=1$ and sample nearest to decision boundary will have $\hat{d_f}(x^i))=0$. Samples having $\hat{d_f}(x^i)$ values close to 0 would generally be more prone to adversarial vulnerability. 

\textbf{Normalised Flipping Frequency}: A higher value of Flipping Frequency $F(x^i)$ represents that the sample is more dependent on high frequency components and hence is more vulnerable (Refer Sec.~\ref{sec:preliminaries}). This is in contrast to DDB metric where higher value of DDB implies less vulnerability. To have a similar effect in case of frequency aspect as well (i.e. higher value representing lower vulnerability), we define a metric named Reversed Normalised Flipping Frequency which can be computed using Eq. \ref{eq:norm_F}).
\begin{equation}\label{eq:norm_F}
   \hat{F}(x^i) = 1-\frac{F(x^i) - min(\tilde{F}(x))}{max(\tilde{F}(x)) - min(\tilde{F}(x))} 
\end{equation}
where $\tilde{F}(x)=\{F(x^1), F(x^2),\cdots,F(x^N)\}$. $\hat{F}(x^i)$ would be close to $0$ for samples which majorly depends on high  frequency components and are more prone to adversarial attacks. Similarly, the value of $\hat{F}(x^i)$ would be $1$ if it is more adversarially robust. 

\textbf{Combined Score}: 
The essence of computing a combined score  $T$ is to capture robustness information from both frequency and DDB. For any sample $x^i$, the combined score $T(x^i)$ can be simply computed using the harmonic mean of Normalised Distance to Decision Boundary $\hat{d}_{f}(x^i)$ and Reversed Normalised Flipping Frequency $\hat{F}(x^i)$ (Eq. \ref{eq:T_score}). Here $\epsilon=10^{-5}$ is a small quantity to ensure numerical stability.  
\begin{equation}\label{eq:T_score}
    T(x^i) = \frac{2 \times \hat{d}_{f}(x^i) \times \hat{F}(x^i)}{\hat{d}_{f}(x^i) + \hat{F}(x^i) + \epsilon}
\end{equation}
The combined score $T$ follows a similar trend compared to other metric, i.e. higher combined score means a lower adversarial vulnerability. We claim this score to be superior than its individual counterparts (verified experimentally in Sec.~\ref{sec:build_trustworthy}) as the harmonic mean is able to captures the combined effects of fluctuations between these two quantities and hence gives a better estimate of adversarial vulnerability. 
\section{Building Trustworthy Systems}
\label{sec:build_trustworthy}
\begin{figure*}[htp]
    \centering
    \includegraphics[width=0.88\linewidth, height=0.33\linewidth]{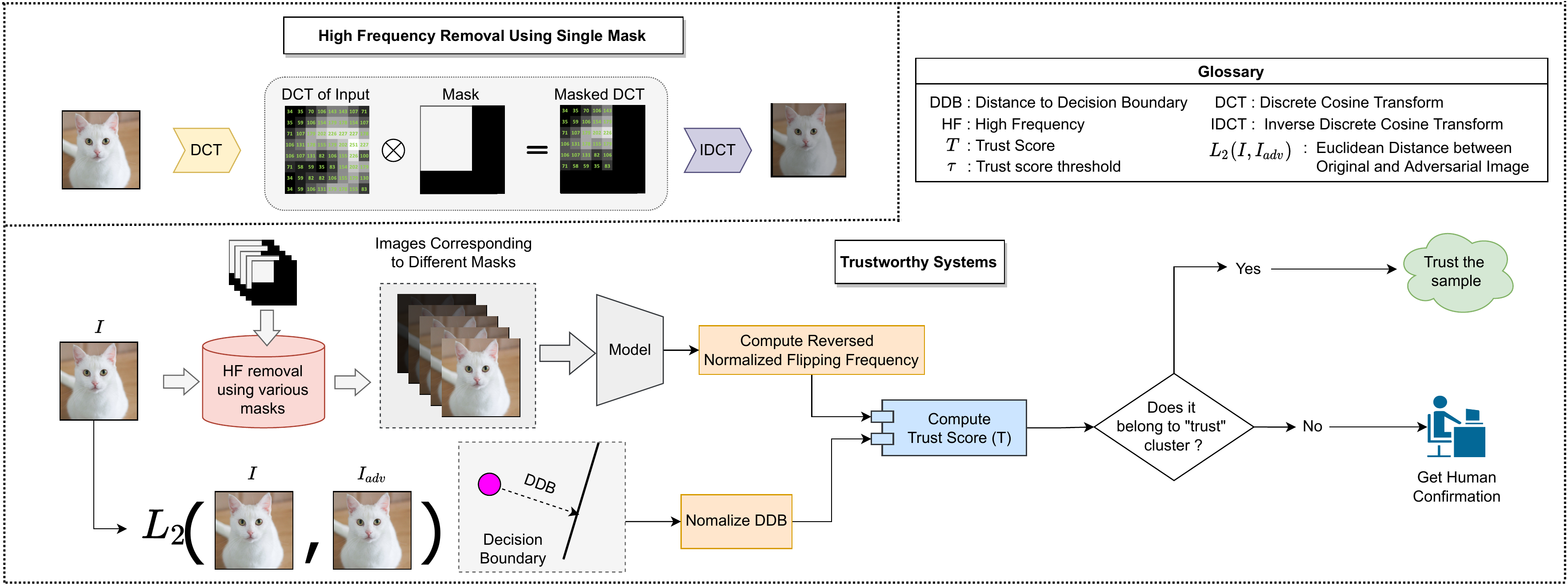}
    \vspace{-0.1in}
    \caption{Detailed steps involved in our proposed design of trustworthy systems. (\textit{Top Row}) Method to remove high frequency components using DCT and IDCT. The image is first transformed to frequency domain using DCT and then a binary mask is applied which selects the required frequencies.
    (\textit{Bottom Row})  For a test sample we compute Reversed Normalized Flipping Frequency (Eq. \ref{eq:norm_F}) and Normalized Distance to Decision to Boundary (Eq. \ref{eq:norm_ddb}). Finally, Trust score (Eq. \ref{eq:T_score}) is computed, based on which the test sample is clustered either into ``Trust cluster" or ``Non-Trust Cluster". Samples that fall into trust cluster are allowed to predict by model, and sample which fall in non-trust cluster are sent to human in form of alert signals for further investigation.}
    \label{fig:trust_system}
\end{figure*}
The combined score as discussed in the previous section is used to quantify the combined effect of both the factors (DDB and frequency). As these factors can better describe adversarial vulnerability of a sample, hence they can be used to determine the trustworthiness of a sample.

High combined score implies high normalized DDB $\hat{d_{f}}$ (i.e. far from decision boundary) and high reverse normalized flipping frequency ($\hat{F}$) (i.e. more dependency on low frequency features), which in turn leads to high trust. Hence combined score $T$ can be treated as \textbf{trust score}. Keeping this in mind, we design a trustworthy system (shown in Figure~\ref{fig:trust_system}) where the model is allowed to only predict when the samples have high trust (i.e. ensuring reliable predictions from the model). If the trust score is low, the system raises an alert signal to human for further investigation. 

To achieve the above objectives i.e. separating out high and low trust samples, we simply use K-means clustering~\cite{1056489} with 2 clusters. The samples with high combined score (trust score) are clustered into the `trust' cluster and samples with low trust score are assigned to `non-trust' cluster. Moreover, this approach allows easy integration with any pretrained model.

Overall, during test time, combined score (trust score) is computed for the test sample using which it gets clustered into one of the clusters (either trust cluster or non-trust cluster). If the test sample belongs to trust cluster, then the model outputs the prediction, otherwise the sample falls into non-trust cluster and alert signal is raised to human domain expert for investigation.



To validate our system,  we perform experiments on CIFAR-10~\cite{krizhevsky2009learning} (containing 10 classes). We conduct experiments by estimating $\hat{d}_{f}$ using different adversarial attacks (i.e. PGD and DeepFool). We also vary the architectures 
for robust (Resnet-18, obtained via adversarial training) and non-robust (Resnet-18 and VGG) settings. 

We also compare our performance (combined multiple perspectives) against the trustworthy system when designed using only one factor (i.e. single perspective of adversarial vulnerability such as DDB). 

We evaluate our method and other baselines using flagging accuracy which is defined as:
\begin{equation}\footnotesize
    \frac{\text{The number of incorrect samples in `non-trust' cluster}}{\text{Total number of samples in `non-trust' cluster}} \times 100
\end{equation}

A higher flagging accuracy is desired. This implies that most of the samples that are sent to humans were wrongly predicted by the model. This can help in reducing human effort and time as human in loop would mostly get those samples which the model can't handle. The results are reported in Table~\ref{tab:trust_scores}. We can observe that T-Score consistently outperforms its individual counterparts by a significant margin across different attacks (PGD and DeepFool) and architectures (Resnet-$18$ and VGG-$11$). Moreover, on an average over different attacks and architectures, our method flags nearly $77\%$ of incorrect predictions which is $\approx10\%$ higher than the average performance observed on normalized DDB.

\begin{table*}[!h]
    \begin{subtable}[h]{0.33\linewidth}
        \renewcommand{\arraystretch}{1.2}
        \centering
        \resizebox{.8\textwidth}{!}{\begin{tabular}{|l|cc|}
            \toprule
            Scores & PGD & DeepFool \\
            \midrule
            $\hat{d}_f$   & 
            23.5 & 32.7\\
            $\hat{F}$   & 29.8 & 29.8 \\
            $T$ (Ours)       & \textbf{35.2} & 
            \textbf{36.2} \\
            \bottomrule
        \end{tabular}%
        }
        \caption{Robust Resnet-18}
    \end{subtable}
    \begin{subtable}[h]{0.33\linewidth}
        \renewcommand{\arraystretch}{1.2}
        \centering
        \resizebox{.8\textwidth}{!}{\begin{tabular}{|l|cc|}
            \toprule
            Scores & PGD & DeepFool \\
            \midrule
            $\hat{d}_f$        & 8.6 & 11.5\\
            $\hat{F}$   & 11.7 & 11.7 \\
            $T$ (Ours)        & 
            \textbf{12.4} & \textbf{13.9} \\
            \bottomrule
        \end{tabular}%
        }
        \caption{ Resnet-18}
    \end{subtable}
    \begin{subtable}[h]{0.33\linewidth}
        \renewcommand{\arraystretch}{1.2}
        \centering
        \resizebox{.8\textwidth}{!}{\begin{tabular}{|l|cc|}
            \toprule
            Scores & PGD & DeepFool \\
            \midrule
            $\hat{d}_f$   & 10.2 & 12.2\\
            $\hat{F}$   & 14.3 & 14.3\\
            $T$ (Ours)        & 
            \textbf{17.6} & \textbf{14.6}\\
            \bottomrule
    \end{tabular}%
    }
    \caption{VGG-11}
    \end{subtable}
    \vspace{-0.1in}
    \caption{Flagging accuracy across various architectures (DDB estimated using PGD and DeepFool adversarial attacks) for Normalised DDB ($\hat{d}_f$), Reversed Normalised Flipping Frequency ($\hat{F}$) and T-score ($T$) on CIFAR10 dataset. It can be observed that T-score outperforms other metrics by a significant margin across various settings.}
    \label{tab:trust_scores}
\end{table*}
\vspace{-0.22in}
\section{Time Efficient Training of Lightweight models using Knowledge Distillation}
\label{sec:KD}
\vspace{-0.07in}
Knowledge Distillation (KD) is a common technique to transfer knowledge from a deep network (called \Te{}) to a shallow network (called \St{}). This is done by matching certain statistics between \Te{} and \St{} networks. The statistics can be logits~\cite{bucilua2006model}, temperature raised softmaxes~\cite{hinton2015distilling}, intermediate feature responses~\cite{romero2014fitnets}, Jacobian~\cite{srinivas2018knowledge} etc. 
It has been popularly used in model compression for classification tasks. The networks after compression become lightweight (less memory footprint) and suitable (less computation and less inference time) to be deployed onto the portable devices.

Most of the existing works use the entire training data to conduct knowledge transfer. As a consequence of this, the training time is quite high. This problem becomes even compounded when large scale training sets are used for training on large architectures. Many times in such situations, sophisticated hardware would be needed for training. Moreover, finding the optimal hyperparameters by running the model several times can become infeasible at times. Therefore, there is a need to do fast training to overcome such issues.

A child quickly learns from just a few examples. Based on this analogy, one way of reducing 
time required is to do training with few selected samples. But if we randomly select those samples, it can result in a significant drop in performance. Therefore, we need to intelligently select the required samples so that we can optimally trade off between training time and distillation accuracy. In other words, our goal is to do \textit{“Quick Training for Fast Knowledge Distillation by only utilizing few training samples”}. There are few works where distillation is performed using few training examples by modifying the \St{}’s architecture~\cite{li2020few} or augmenting pseudo data with few training samples~\cite{kimura2019few}. But unlike these works, we do not aim to modify \St{}’s architecture or generate pseudo samples which may bring good performance at the cost of large training time. 

\begin{equation}\small
    \mathcal{L}=\sum_{(x,y) \in \mathbb{\hat{D}} }(1-\lambda) \mathcal{L}_{kd}(St(x;\theta_S,\tau),Te(x;\theta_T,\tau))+\lambda  \mathcal{L}_{ce}({\hat{y}}_S, y)
    \label{eqn:kd}
\end{equation}
Let the optimal transfer set ($\hat{D}$) contain the selected training samples which is $k\%$ of entire training data such that the cardinality of entire dataset ($D$) would be much greater than the cardinality of
$\hat{D}$ i.e. $|D|$\(\gg\)$|\hat{D}| = k\%$ of $D$. The distillation can then be performed on samples of $\hat{D}$ as in eq.~\ref{eqn:kd}. Here, $Te$ and $St$ denotes \Te{} and \St{} networks with parameters $\theta_T$ and $\theta_S$ respectively. We use a robust ResNet-18 network as our \Te{} model ($Te$) and MobileNet-V2 as our \St{}  network ($St$). $L_{kd}$ represents distillation loss while cross entropy is denoted by $L_{ce}$. The temperature ($\tau$) and $\lambda$ are hyperparameters.  

In order to carefully select only a subset of training samples and use them as a transfer set to conduct the distillation, we take into account the factors discussed in previous subsections i.e. sample's distance to decision boundary and model's frequency reliance on that sample 
for its prediction. 
For a limited data budget, we rank all the samples in the training data according to our trust-score metric (described in Sec~\ref{sec:build_trustworthy} ; with respect to the 
teacher model) and select a fixed amount of most-trustworthy samples for each class. The student model is then trained on those trust-worthy samples directly using standard distillation losses (described in Eq.~\ref{eqn:kd}). We adopt a (uniform) random sample selection strategy as our baseline, where we randomly select a fixed number of samples from each class. Furthermore, 
the samples lying close to the teacher model’s decision boundary are often important for the student network to 
successfully mimic the teacher’s decision boundary \cite{heo2019knowledge}. Hence,  we also report the student model's performance when the samples closest (for each class separately) to the decision boundary are selected. We perform experiments with multiple sample selection budgets (i.e. $100$, $120$ and $150$ samples per class).
\begin{figure}[b]
    \centering
    \includegraphics[width=\linewidth]{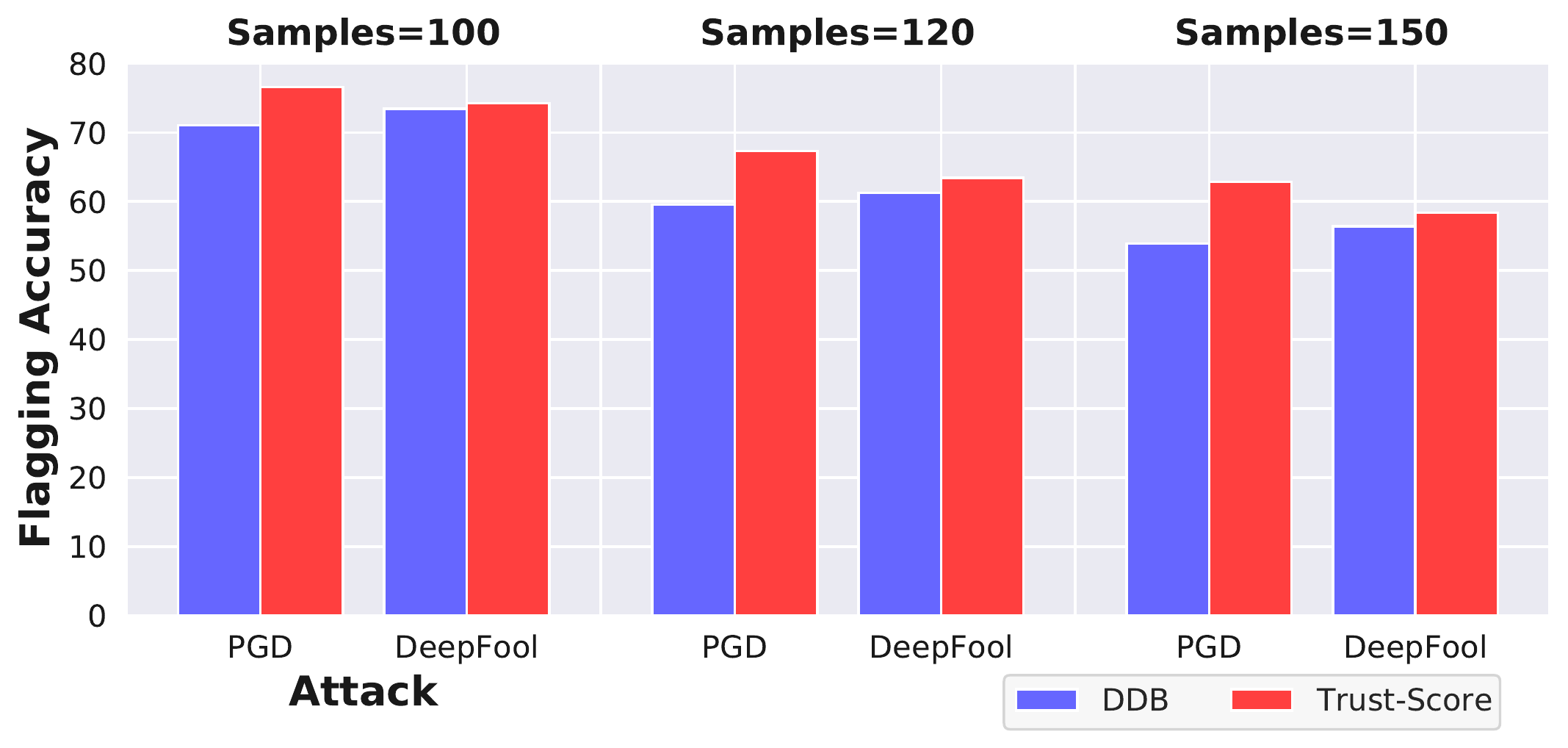}
    \vspace{-0.2in}
    \caption{Flagging accuracy comparison for (Normalized) DDB score v/s (Our Proposed) Trust-Score (flagging) metrics across various attacks (PGD and DeepFool) and multiple per-class sample-selection budgets ($100$, $120$, $150$) for the model trained via our Trust Score based sample selection strategy.}
    \label{fig:bar_graph}
\end{figure}

In Table~\ref{tab:my-table} we observe that our proposed trust-worthy sample selection strategy consistently outperforms both the random baseline and the conventional distance to decision boundary 
across multiple sample selection budgets. In Figure~\ref{fig:bar_graph} we further analyze the flagging accuracy performance (normalized DDB and our trust score metrics) for the best performing \St{} models (i.e. models distilled via our trust-worthy sample selection strategy) across different sample-budgets. We observe that even on a specific application (KD in this case), our flagging accuracy using trust score is consistently better (or atleast similar) 
than the performance on normalized DDB. Apart from this, even the models trained via suboptimal sample selection strategies (such as random and closest to DDB), we obtained better performance. For instance, we notice an improvement in flagging accuracy from $74.0\%$ to $75.83\%$ on random baseline and $73.41\%$ to $75.94\%$ on closest to DDB (estimated using PGD) with a budget of $100$ samples per class. Similar trend is followed across other sample budgets.

\begin{table}[htp]
\centering
\scalebox{0.85}{
\begin{tabular}{|c|c|c|c|}
\hline
\textbf{\begin{tabular}[c]{@{}c@{}}Samples Selection\\ (Per Class)\end{tabular}} &
  \textbf{\begin{tabular}[c]{@{}c@{}}Sample Selection\\ Criteria\end{tabular}} &
  \textbf{\begin{tabular}[c]{@{}c@{}}$\tau=30.0$\\ $\lambda=0.2$\end{tabular}} &
  \textbf{\begin{tabular}[c]{@{}c@{}}$\tau=8.0$\\ $\lambda=0.2$\end{tabular}} \\ \hline
\multirow{3}{*}{100} & Random                                                                   & 27.06 & 25.42 \\ \cline{2-4} 
                     & \begin{tabular}[c]{@{}c@{}}Closest to DDB \\ (Conventional)\end{tabular} & 24.19 & 23.4  \\ \cline{2-4} 
                     & \begin{tabular}[c]{@{}c@{}}\textbf{Ours} \\ (Trust Score based)\end{tabular}      & \textbf{41.63} & \textbf{36.65} \\ \hline \hline
\multirow{3}{*}{120} & Random                                                                   & 37.48 & 22.98 \\ \cline{2-4} 
                     & \begin{tabular}[c]{@{}c@{}}Closest to DDB \\ (Conventional)\end{tabular} & 24.34 & 20.90 \\ \cline{2-4} 
                     & \begin{tabular}[c]{@{}c@{}}\textbf{Ours} \\ (Trust Score based)\end{tabular}      & \textbf{44.64} & \textbf{44.52} \\ \hline \hline
\multirow{3}{*}{150} & Random                                                                   & 46.10 & 41.59 \\ \cline{2-4} 
                     & \begin{tabular}[c]{@{}c@{}}Closest to DDB \\ (Conventional)\end{tabular} & 22.87 & 24.92 \\ \cline{2-4} 
                     & \begin{tabular}[c]{@{}c@{}}\textbf{Ours} \\ (Trust Score based)\end{tabular}      & \textbf{50.38} & \textbf{50.73} \\ \hline
\end{tabular}
}
\vspace{-0.1in}
\caption{Performance of \St{} (MobileNet-V2) when different budgets of samples (per-class) are selected, using different sample selection strategies on CIFAR-10 dataset.}
\label{tab:my-table}
\end{table}
\vspace{-0.15in}
\vspace{-0.1in}
\section{Conclusion}
We presented the traditional factor for quantifying adversarial vulnerability (i.e. distance to decision boundary) and discussed its limitations (i.e. samples having similar distance from decision boundary can have different vulnerabilities). To overcome this, we proposed a holistic view by considering other factors such as model reliance on high frequency features for its prediction. We combined the multiple perspectives to propose a holistic metric that can be used to quantify sample trust. Hence, we designed trustworthy systems using the holistic metric that yields reliable predictions and provides alert signals to human domain-expert whenever encountering non-trustable samples with better precision than individual factors considered alone. We also showed the utility of multiple factors that  characterise adversarial vulnerability in a knowledge distillation setup, where the student network can be trained efficiently in a limited data-setting using the samples selected based on the vulnerability factors. In the future, we plan to extend this work by exploring other aspects of adversarial vulnerability beyond distance to decision boundary and high-frequency reliance, and investigating their utility across different applications.

{\small
\bibliographystyle{ieee_fullname}
\bibliography{references}
}

\end{document}